\documentclass{article}

\usepackage{microtype}
\usepackage{booktabs} 

\usepackage{hyperref}
\usepackage{url}
\usepackage{amsthm}
\usepackage{amsfonts}
\usepackage[dvipsnames]{xcolor}
\usepackage{lipsum}  
\usepackage{multicol}
\usepackage{tikz}
\usepackage{lipsum}
\usepackage{array}

\usepackage{amsmath}
\usepackage{amssymb}
\usepackage{mathrsfs}
\usepackage{graphicx}
\usepackage{subfig}
\usepackage{floatrow}
\usepackage{algorithm}
\usepackage{algorithmic}
\usepackage{float}

\newcommand{\gb}{\boldsymbol{g}}

\newcommand{\mb}{\boldsymbol{m}}

\newcommand{\ub}{\boldsymbol{u}}

\newcommand{\xb}{\boldsymbol{x}}
\newcommand{\yb}{\boldsymbol{y}}

\newcommand{\alphab}{\boldsymbol{\alpha}}

\newcommand{\E}{\mathrm{E}}

\DeclareMathOperator*{\argmin}{arg\,min}
\DeclareMathOperator*{\argmax}{arg\,max}

\usepackage{hyperref}

\usepackage[accepted]{icml2019}

\icmltitlerunning{Concrete Autoencoders}

\begin{document}

\tikzset{%
  every neuron/.style={
    circle,
    draw,
    minimum size=1cm
  },
  neuron missing/.style={
    draw=none, 
    scale=3,
    text height=0.333cm,
    execute at begin node=\color{black}$\vdots$
  },
}

\twocolumn[
\icmltitle{Concrete Autoencoders for Differentiable Feature Selection and Reconstruction}



\icmlsetsymbol{equal}{*}

\begin{icmlauthorlist}
\icmlauthor{Abubakar Abid}{equal,stanfordee}
\icmlauthor{Muhammad Fatih Balin}{equal,turkey}
\icmlauthor{James Zou}{stanfordbds}
\end{icmlauthorlist}

\icmlaffiliation{stanfordee}{Department of Electrical Engineering, Stanford University, Stanford, United States}
\icmlaffiliation{turkey}{Bogazici University, Istanbul, Turkey}
\icmlaffiliation{stanfordbds}{Department of Biomedical Data Sciences, Stanford University, Stanford, United States}

\icmlcorrespondingauthor{James Zou}{jamesz@stanford.edu}

\icmlkeywords{Machine Learning, ICML}

\vskip 0.3in
]



\printAffiliationsAndNotice{\icmlEqualContribution} 

\begin{abstract}
We introduce the \textit{concrete autoencoder}, an end-to-end differentiable method for global feature selection, which efficiently identifies a subset of the most informative features and simultaneously learns a neural network to reconstruct the input data from the selected features. Our method is unsupervised, and is based on using a concrete selector layer as the encoder and using a standard neural network as the decoder. During the training phase, the temperature of the concrete selector layer is gradually decreased, which encourages a user-specified number of discrete features to be learned. During test time, the selected features can be used with the decoder network to reconstruct the remaining input features.  We evaluate concrete autoencoders on a variety of datasets, where they significantly outperform state-of-the-art methods for feature selection and data reconstruction. In particular, on a large-scale gene expression dataset, the concrete autoencoder selects a small subset of genes whose expression levels can be use to impute the expression levels of the remaining genes. In doing so, it improves on the current widely-used expert-curated L1000 landmark genes, potentially reducing measurement costs by 20\%.  The concrete autoencoder can be implemented by adding just a few lines of code to a standard autoencoder.
\end{abstract}

\section{Introduction}

High-dimensional datasets often pose a challenge for machine learning algorithms. Feature selection methods aim to reduce dimensionality of data by identifying the subset of relevant features in a dataset. A large number of  algorithms have been proposed for feature selection in both supervised and unsupervised settings \citep{kohavi1997wrappers, wang2013feature}. These methods provide insight into the relationship between features in complex data and can simplify the process of training downstream models. Feature selection is particularly useful if the data with the full set of features is expensive or difficult to collect, as it can eliminate the need to measure irrelevant or redundant features.

As a motivating example, consider a dataset that consists of the expression of various genes across tissue samples. Such ``omics'' measurements are increasingly carried out to fully characterize biological samples at the individual and single-cell level \citep{bock2016multi, huang2017more}. Yet it remains  expensive to conduct all of the biological assays that are needed to characterize such samples. It is natural to ask: \textit{what are the most important features in this dataset? Are there redundant features that do not need to be measured?} The idea of only measuring a subset of biological features and then reconstructing the remaining features is not new; in fact, this line of thinking has motivated the identification of the landmark genes, also known as the L1000 genes, which are a small subset of the over 20,000 human genes. The expression levels of the L1000 are strongly correlated with the expression levels of other genes, and thus this subset can be measured cheaply and then used to impute the remaining gene expression levels \citep{lamb2006connectivity}.


\begin{figure*}[!htb]
\centering
\subfloat[]{\includegraphics[width=0.24\linewidth]{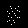}} \;
\subfloat[]{\includegraphics[width=0.24\linewidth]{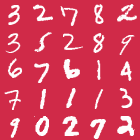}} \;
\subfloat[]{\includegraphics[width=0.24\linewidth]{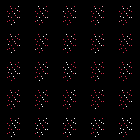}} \;
\subfloat[]{\includegraphics[width=0.24\linewidth]{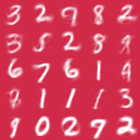}} 
\vspace{-0.3cm}
\caption{\textbf{Demonstrating concrete autoencoders on the MNIST dataset.} Here, we show the results of using concrete autoencoders to select in an unsupervised manner the $k=20$ most informative pixels of images in the MNIST dataset. (a) The 20 selected features (out of the 784 pixels) on the MNIST dataset are shown in white. (b) A sample of input images in MNIST dataset with the top 2 rows being training images and the bottom 3 rows being test images. (c) The same input images with only the selected features shown as colored dots. (d). The reconstructed versions of the images, using only the 20 selected pixels, shows that generally the digit is identified correctly and some stylistic features, such as the swirl in the digit ``2'', are captured. (\textit{cf.} figures in Appendix \ref{appendix:single-classes} which show the results of applying concrete autoencoder to individual classes of digits.) } \label{fig:mnist-full}
\end{figure*}

The problem of feature selection is different from the more general problem of dimensionality reduction. Standard techniques for dimensionality reduction, such as principal components analysis \citep{hotelling1933analysis} and autoencoders \citep{hinton2006reducing}, are be able to represent data with fewer dimensions while preserving maximal variance or minimizing reconstruction loss. However, such methods do not select a set of features present in the original dataset, and thus cannot be directly used to eliminate redundant features and reduce experimental costs. 
We emphasize the \textit{unsupervised} nature of this problem: specific prediction tasks may not be known ahead of time, and thus it is important to develop methods that can identify a subset of features while allowing imputation of the remaining features with minimal distortion for arbitrary downstream tasks.

In this paper, we propose a new end-to-end method to perform feature subset selection and imputation that leverages the power of deep autoencoders for discrete feature selection. Our method, the \textit{concrete autoencoder}, uses a relaxation of the discrete distribution, the Concrete distribution \citep{maddison2016concrete}, and the reparametrization trick \citep{kingma2013auto} to differentiate through an arbitrary (e.g. reconstruction) loss and select input features to minimize this loss. A visual example of results from our method is shown in Fig. \ref{fig:mnist-full}, where the concrete autoencoder selects the 20 most informative pixels (out of a total of 784) on the MNIST dataset, and reconstructs the original images with high accuracy. 

We test concrete autoencoders on a variety of datasets, and find that they generally outperform state-of-the-art methods for feature selection and data reconstruction. We have made the code for our algorithm and experiments available on a public repository\footnote{Code available at: \url{https://github.com/mfbalin/Concrete-Autoencoders}}.

\paragraph{Related Works}
Feature selection methods are generally divided into three classes: filter, wrapper, and embedded methods. Filter methods rank the features of the dataset using statistical tests, such as the variance, in order to select features that individually maximize the desired criteria \citep{battiti1994using, duda2012pattern}. Filter methods generally do not consider interactions between features and do not provide a way to impute the remaining features; one must train a separate algorithm to do so. Wrapper methods select subsets of features that maximize a objective function, which is optimized over the choice of input features using a black-box optimization method, such as sequential search or genetic algorithms \citep{kohavi1997wrappers, goldberg1988genetic}. Since wrapper methods evaluate subsets of features, they are able to detect potential relationships between features, but usually at the expense of increased computation time. Embedded methods also consider relationships between features but generally do so more efficiently as they incorporate feature selection into the learning phase of another algorithm. A well-known example is the Lasso \citep{tibshirani1996regression}, which can be used to select features for regression by varying the strength of $\ell_1$ regularization. 

Many embedded unsupervised feature selection algorithms use regularization as the means to select discrete features. The popular UDFS algorithm \citet{yang2011l2} incorporates $\ell_{2,1}$ regularization on a set of weights applied to the input to select features most useful for local discriminative analysis. Similarly, the MCFS algorithm \citep{cai2010unsupervised} uses regularization to solve for the features which preserve the clustering structure in the data. The recently-proposed AEFS \citep{han2017autoencoder} algorithm  also uses $\ell_{2,1}$ regularization on the weights of the encoder that maps the input data to a latent space and optimizes these weights for their ability to reconstruct the original input. 

In this paper, we select discrete features using an embedded method but without resorting to  regularization. Rather, we use a relaxation of the discrete random variables, the Concrete distribution \citep{maddison2016concrete}, which allows a low-variance estimate of the gradient through discrete stochastic nodes. By using Concrete random variables, we can directly parametrize the selection of input features, and differentiate through the parameters. As we show through experiments in Section \ref{section:experiments}, this leads to lower reconstruction errors on real-world datasets compared to the aforementioned regularization-based methods. 


\section{Problem Formulation}
We now describe the problem of global feature selection. Although global feature selection is relevant for both unsupervised and supervised settings, we describe here the unsupervised case, which is the primary focus of this paper, and defer discussion of the supervised case to Appendix \ref{appendix:supervised}.

Consider a data-generating probability distribution $p(\xb)$ over a $d$-dimensional space. The goal is to learn a subset $S \subseteq \{1, 2  \ldots d\}$ of features of specified size $\lvert S \rvert = k$ and also learn a reconstruction function $f_\theta(\cdot):\mathbb{R}^k \to \mathbb{R}^d$, such that the expected loss between the reconstructed sample $f_\theta(\xb_S)$ and the original sample $\xb$ is minimized, where $\xb_S \in \mathbb{R}^k$ consists of those elements $\xb_i$ such that $i \in S$. In other words, we would like to optimize
\begin{align}
\argmin_{S, \theta} \E_{p(x)}[\lVert f_\theta(\xb_S) - \xb \lVert_2] 
\label{eq:objective}
\end{align}
In practice, we do not know $p(\xb)$; rather we have $n$ samples, generally assumed to be drawn i.i.d. from $p(\xb)$. These samples can be represented in a data matrix $X \in \mathbb{R}^{n \times d}$, and so the goal becomes choosing $k$ columns of $X$ such that sub-matrix $X_S \in \mathbb{R}^{n \times k}$, defined analogously to $\xb_S$, can be used to reconstruct the original matrix $X$. Let us overload $f_\theta(X_S)$ to mean the matrix that results from applying $f_\theta(\cdot)$ to each of the rows of $X_S$ and stacking the resulting outputs. We seek to minimize the empirical reconstruction error:
\begin{align}
\argmin_{S, \theta} \lVert f_\theta(X_S) - X \lVert_F,
\label{eq:objective2}
\end{align}
where $\lVert \cdot \rVert_F$ denotes the Frobenius norm of the matrix. The principal difficulty in solving (\ref{eq:objective2}) is the optimization over the discrete set of features $S$, whose choices grow exponentially in $d$. Thus, even for simple choices of $f_\theta(\cdot)$, such as linear regression, the optimization problem in (\ref{eq:objective2}) is NP-hard to solve \cite{amaldi1998approximability}.

Furthermore, the complexity of $f_\theta(\cdot)$ can significantly affect reconstruction error and even the choice of $S$. More expressive and non-linear choices for $f_\theta(\cdot)$, such as neural networks, will naturally allow for lower reconstruction error, potentially at the expense of a more difficult optimization problem. We seek to develop a method that can approximate the solution for any given class of functions $f_\theta(\cdot)$, from linear regression to deep fully-connected neural networks. 

Finally, we note that the choice of mean-squared reconstruction error as the metric for optimization in (\ref{eq:objective}) and  (\ref{eq:objective2}) stems from the fact that it is a smooth differentiable function that serves as a proxy for many downstream analyses, such as clustering performance and classification accuracy. However, other differentiable metrics, such as a variational approximation of the mutual information between $f_\theta(X_S)$ and $X$, may be considered as well \citep{pmlr-v80-chen18j}.

\section{Proposed Method}
\label{section:method}

\begin{figure*}[!htb]
\centering
\subfloat[]{
\resizebox{9.5cm}{!}{
\raisebox{0.5cm}{\begin{tikzpicture}[x=1.5cm, y=1.5cm, >=stealth]

\draw[rounded corners, ultra thick, draw=black, dashed, fill=gray, fill opacity=0.05, draw opacity=0.15] (-1.3, -3.1) rectangle (2.7, 2.3) {};
\draw[rounded corners, ultra thick, draw=black, dashed, fill=gray, fill opacity=0.05, draw opacity=0.15] (3.1, -3.1) rectangle (7.5, 2.3) {};

\foreach \m/\l [count=\y] in {1,2,3,missing,4}
  \node [every neuron/.try, neuron \m/.try] (input-\m) at (0,2.5-\y) {};


\foreach \m [count=\y] in {1,missing,2}
  \node [every neuron/.try, neuron \m/.try ] (hidden-\m) at (2,2-\y*1.25) {};

\foreach \m [count=\y] in {1,missing,2}
  \node [every neuron/.try, neuron \m/.try ] (hidden2-\m) at (4,1.5-\y) {};
  
\foreach \m/\l [count=\y] in {1,2,3,missing,4}
  \node [every neuron/.try, neuron \m/.try] (output-\m) at (6,2.5-\y) {};

\foreach \l [count=\i] in {1,2,3,d}
  \draw [<-] (input-\i) -- ++(-1,0)
    node [above, midway] {$\xb_\l$};

\foreach \l [count=\i] in {1,k}
  \node [above] at (hidden-\i.north) {$u^{(\l)}$};

\foreach \l [count=\i] in {1,2,3,d}
  \draw [->] (output-\i) -- ++(1,0)
    node [above, midway] {$\hat{\xb}_\l$};

\foreach \i in {1,...,4}
  \foreach \j in {1,...,2}
    \draw [->, color=brown] (input-\i) -- (hidden-\j);

\foreach \i in {1,...,2}
  \foreach \j in {1,...,2}
    \draw [->, color=teal] (hidden-\i) -- (hidden2-\j);

\foreach \i in {1,...,4}
  \foreach \j in {1,...,2}
    \draw [->, color=teal] (hidden2-\j) -- (output-\i);

  \node [align=right, above, color=brown, font=\bfseries] at (1.35, 1.9) {Concrete selector layer};
  \node [align=left, above, color=teal, font=\bfseries] at (4, 1.85) {Decoder layers};
  \node [align=left, above, color=teal, ] at (3.95, 1.45) {Arbitrary $f_\theta(\cdot)$};

\end{tikzpicture}}}

} \;
\subfloat[]{\includegraphics[width=0.36\linewidth]{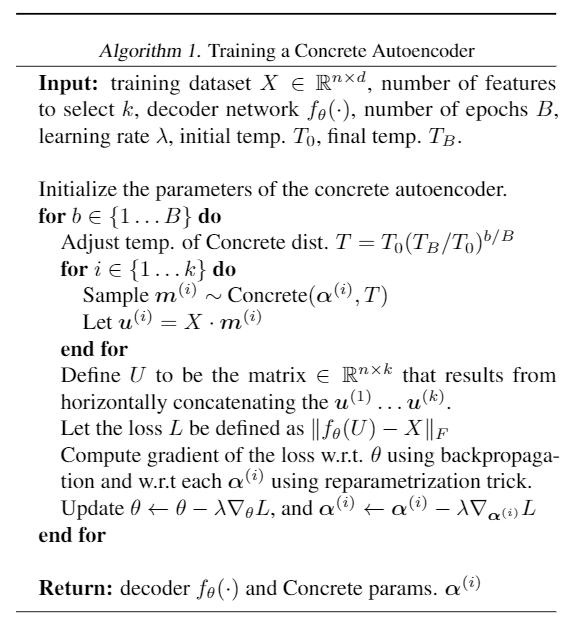}} 
\vspace{-0.3cm}
\caption{\textbf{Concrete autoencoder architecture and pseudocode.} (a) The architecture of a concrete autoencoder consists of a single encoding layer, shown in \textcolor{brown}{brown}, and arbitrary decoding layers (e.g. a deep feedforward neural network), shown in \textcolor{teal}{teal}. The encoder has one neuron for each feature to be selected. During the training phase, the $i^\mathrm{th}$ neuron $u^{(i)}$ takes the value $\xb^\top \mb^{(i)}, \; \mb^{(i)} \sim  $ Concrete$(\alphab^{(i)}, \mathrm{T})$. During test time, these weights are fixed and the element with the highest value in $\alphab^{(i)}$ is selected by the corresponding $i^\mathrm{th}$ hidden neuron. The architecture of the \textit{decoder} remains the same during train and test time, namely that $\hat{\xb} = f_\theta(\ub)$, where $\ub$ is the vector consisting of each $u^{(i)}$. (b) Here, we show pseudocode for the concrete autoencoder algorithm, see Appendix \ref{appendix:pseudocode} for more details.} \label{fig:methods}
\end{figure*}
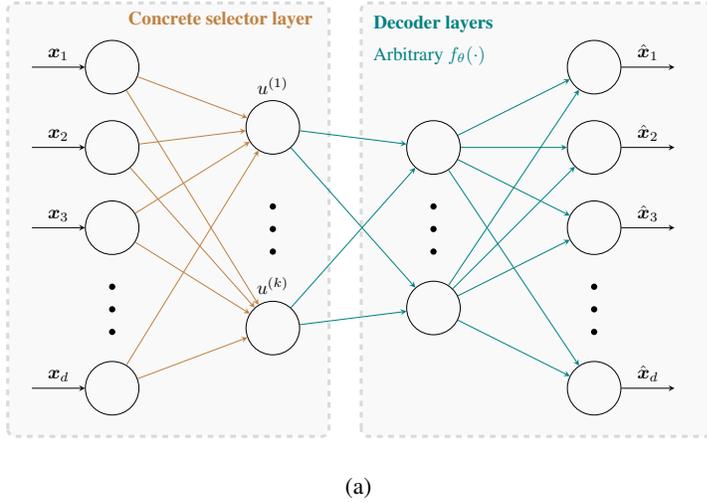

The concrete autoencoder is an adaption of the standard autoencoder \citep{hinton2006reducing} for discrete feature selection. Instead of using series of fully-connected layers for the encoder, we propose a \textit{concrete selector layer} with a user-specified number of nodes, $k$. This layer selects stochastic linear combinations of input features during training, which converge to a discrete set of $k$ features by the end of training and during test time. 

The way in which input features are combined depends on the \textit{temperature} of this layer, which we modulate using a simple annealing schedule. As the temperature of the layer approaches zero, the layer selects $k$ individual input features. The decoder of a concrete autoencoder, which serves as the reconstruction function, is the same as that of a standard autoencoder: a neural network whose architecture can be set by the user based on dataset size and complexity. In effect, then, the concrete autoencoder is a method for selecting a discrete set of features that are optimized for an arbitrarily-complex reconstruction function. We describe the ingredients for our method in more detail in the next two subsections.

\subsection{Concrete Selector Layer}

The concrete selector layer is based on Concrete random variables \citep{maddison2016concrete, jang2016categorical}. A Concrete random variable can be sampled to produce a continuous relaxation of the one-hot vector. The extent to which the one-hot vector is relaxed is controlled by a temperature parameter $T \in (0, \infty)$. To sample a Concrete random variable in $d$ dimensions with parameters $\alphab \in \mathbb{R}_{>0}^d$ and $T$, one first samples a $d$-dimensional vector of i.i.d. samples from a Gumbel distribution \citep{gumbel1954statistical}, $\gb$. Then each element of the sample $\mb$ from the Concrete distribution is defined as:
\begin{align}
\mb_j = \frac{\exp((\log \alphab_j + \gb_j)/T)}{\sum_{k=1}^d \exp((\log \alphab_k + \gb_k)/T)},
\end{align}
where $\mb_j$ refers to the $j^\text{th}$ element in a particular sample vector. In the limit $T \to 0$, the concrete random variable smoothly approaches the discrete distribution, outputting one hot vectors with $\mb_j = 1$ with probability $\alphab_j / \sum_p \alphab_p$. The desirable aspect of the Concrete random variable is that it allows for differentiation with respect to its parameters $\alphab$ via the reparametrization trick \citep{kingma2013auto}.

We use Concrete random variables to select input features in the following way. For each of the $k$ nodes in the concrete selector layer, we sample a $d$-dimensional Concrete random variable $\mb^{(i)}, i \in \{1 \ldots k\}$ (note that the superscript here indexes the node in the selector layer, whereas the subscript earlier referred to the element in the vector). The $i^\text{th}$ node in the selector layer $u^{(i)}$ outputs $\xb \cdot \mb^{(i)}$. This is, in general, a weighted linear combination of the input features, but notice that when $T \to 0$, each node in the concrete selector layer outputs exactly one of the input features. After the network is trained, during test time, we thus replace the concrete selector layer with a discrete $\arg \max$ layer in which the output of the $i^\text{th}$ neuron is $\xb_{\arg \max_j {\alphab^{(i)}_j}}$. 

We randomly initialize $\alphab_i$ to small positive values,
to encourage the selector layer to stochastically explore different linear combinations of input features. However, as the network is trained, the values of $\alphab_i$ become more sparse,
as the network becomes more confident in particular choices of input features, reducing the stochasticity in selected features. The concrete autoencoder architecture is shown in Fig. \ref{fig:methods}(a) and the pseudocode for training in Fig. \ref{fig:methods}(b). 

\subsection{Annealing Schedule}

The temperature of Concrete random variables in the concrete selector layer has a significant affect on the output of the nodes. If the temperature is held high, the concrete selector layer continuously outputs a linear combination of features. On the contrary, if the temperature is held low, the concrete selector layer is not able to explore different combinations of features and converges to a poor local minimum. Neither fixed temperature allows the concrete selector layer to converge to informative features. 

\begin{figure}[!htb]
\centering
\subfloat{\includegraphics[width=\linewidth]{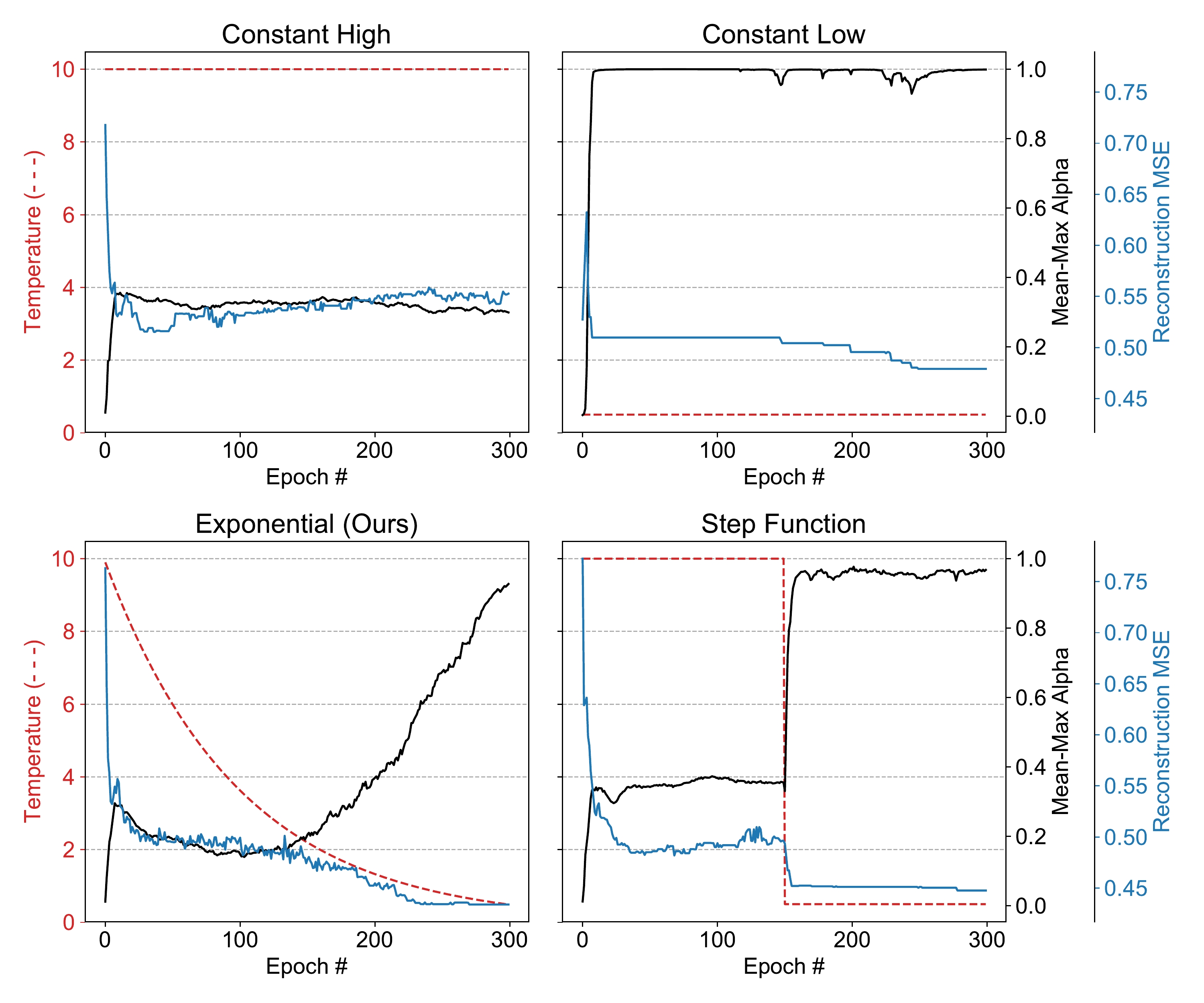}} 
\vspace{-0.3cm}
\caption{\textbf{Annealing schedules for the concrete autoencoder.} Here, we show the effect of different annealing schedules on a concrete autoencoder trained on the MNIST dataset with $k=20$ selected features. At each epoch, we plot the temperature in \textcolor{red}{red}, average of the largest value in each concrete sample $\mb^{(i)}$  in black, as well the reconstruction error (using linear regression with the top $k=20$ features on validation data), shown in \textcolor{blue}{blue}. If the temperature is kept high, the concrete samples do not converge to individual features, and the reconstruction error remains large (top left). If the temperature is kept low, the samples immediately converge to poor features, and the error remains large (top right). If the temperature is exponentially decayed (the annealing schedule we use), the samples converge to informative features, and the reconstruction error reaches a suitable minimum (bottom left). Finally, if the temperature is dropped abruptly, the samples converge, but the error is suboptimal (bottom right).} \label{fig:annealing_schedules}
\end{figure}

Instead, we propose a simple annealing schedule that sets the temperature for all of the concrete variables, initially beginning with a  high temperature $T_0$ and gradually decaying the temperature until a final temperate $T_B$ at each epoch according to a first-order exponential decay:
     $  T(b) = T_0 (T_B / T_0) ^{b/B}$
where $T(b)$ is the temperature at epoch number $b$, and $B$ is the total number of epochs. We compare various methods for setting the temperature of the concrete selector nodes in Fig. \ref{fig:annealing_schedules}. We find that this annealing schedule allows the concrete selector layer to effectively stochastically explore combinations of features in the initial phases of training, while in the later stages of training, the lowered temperature allows the the network to converge to informative individual features.

 

\section{Experiments}
\label{section:experiments}

In this section, we carry out experiments to compare the performance of concrete autoencoders to other feature subset selections on standard public datasets. For all of the experiments, we use Adam optimizer with a learning rate of $10^{-3}$. The initial temperature of the concrete autoencoder $T_0$ was set to $10$ and the final temperature $T_B$ to $0.01$. We trained the concrete autoencoder until until the mean of the concrete samples exceeded 0.99. For UDFS and AEFS, we swept values of each regularization hyperparameter
and report the results with optimal hyperparameters according to mean squared error for reconstruction for each method.

Furthermore, since the reconstruction $f_\theta(\cdot)$ can overfit to patterns particular to the training set, we divide each dataset randomly into train, validation, and test datasets  according to a 72-8-20 split\footnote{For the MNIST, MNIST-Fashion, and Epileptic datasets, we only used 6000, 6000 and 8000 samples respectively to train and validate the model (using a 90-10 split), because of long runtime of the UDFS algorithm. The remaining samples were used for the test set.}. We use the training set to learn the parameters of the concrete autoencoders, the validation set to select optimal hyperparameters, and the test set to evaluate generalization performance, which we report below.

We compare concrete autoencoders to many of the unsupervised feature selection methods mentioned in Related Works including UDFS, MCFS, and AEFS. We also include principal feature analysis (PFA), proposed by \citet{lu2007feature}, which is a popular method for selecting discrete features based on PCA, as well as a spectral method, the Laplacian score \citep{he2006laplacian}. Where available, we made use of \texttt{scikit-feature} implementation of each method \citep{li2016feature}. In our experiments, we also include, as upper bounds on performance, dimensionality reduction methods that are not restricted to choosing individual features. In experiments with linear decoders, we use PCA and in experiments with non-linear decoders, we use equivalent autoencoders. The methods, because they allow $k$ \textit{combinations} of features, bound the performance of any feature selection techniqe.

\setcounter{figure}{1} 

\begin{figure*}[!htb]
\centering
\floatbox[{\capbeside\thisfloatsetup{capbesideposition={right,top},capbesidewidth=6.5cm}}]{figure}[\FBwidth]
{\caption{\textbf{Results on the ISOLET dataset using non-linear decoders.} Here, we compare the concrete autoencoder (CAE) to other feature selection methods using a 1-hidden layer neural network as the reconstructor. (a) We find that across all values of $k$ tested, concrete autoencoders have lowest reconstruction errors (b) We find that the features learned by the concrete autoencoder also tend result in higher classification accuracies.}\label{fig:results_nonlinear}}{
\subfloat[]{\includegraphics[width=0.54\linewidth, trim={0.4cm 0 0.4cm 1cm},clip]{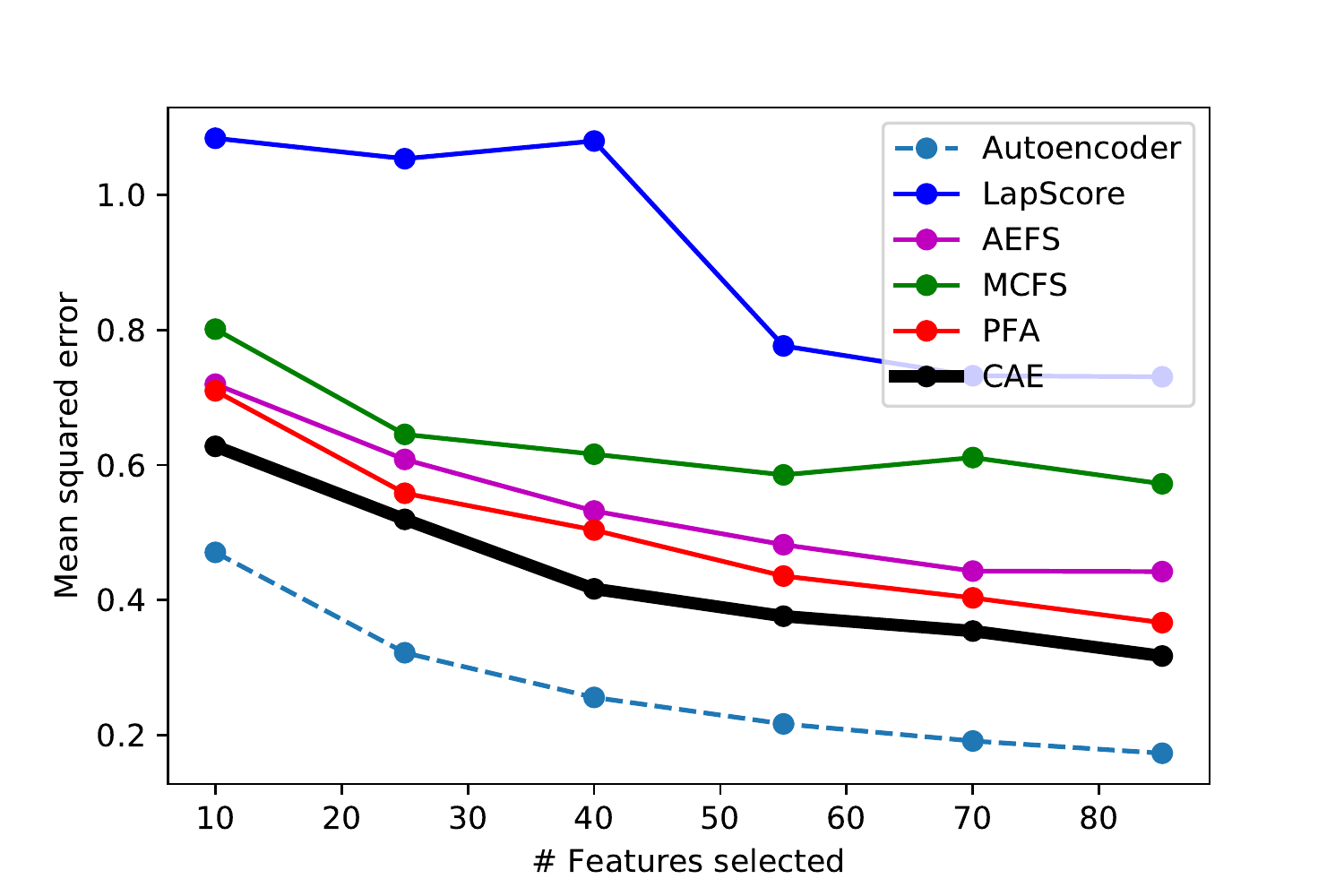}}
\subfloat[]{\includegraphics[width=0.54\linewidth, trim={0.4cm 0 0.4cm 1cm},clip]{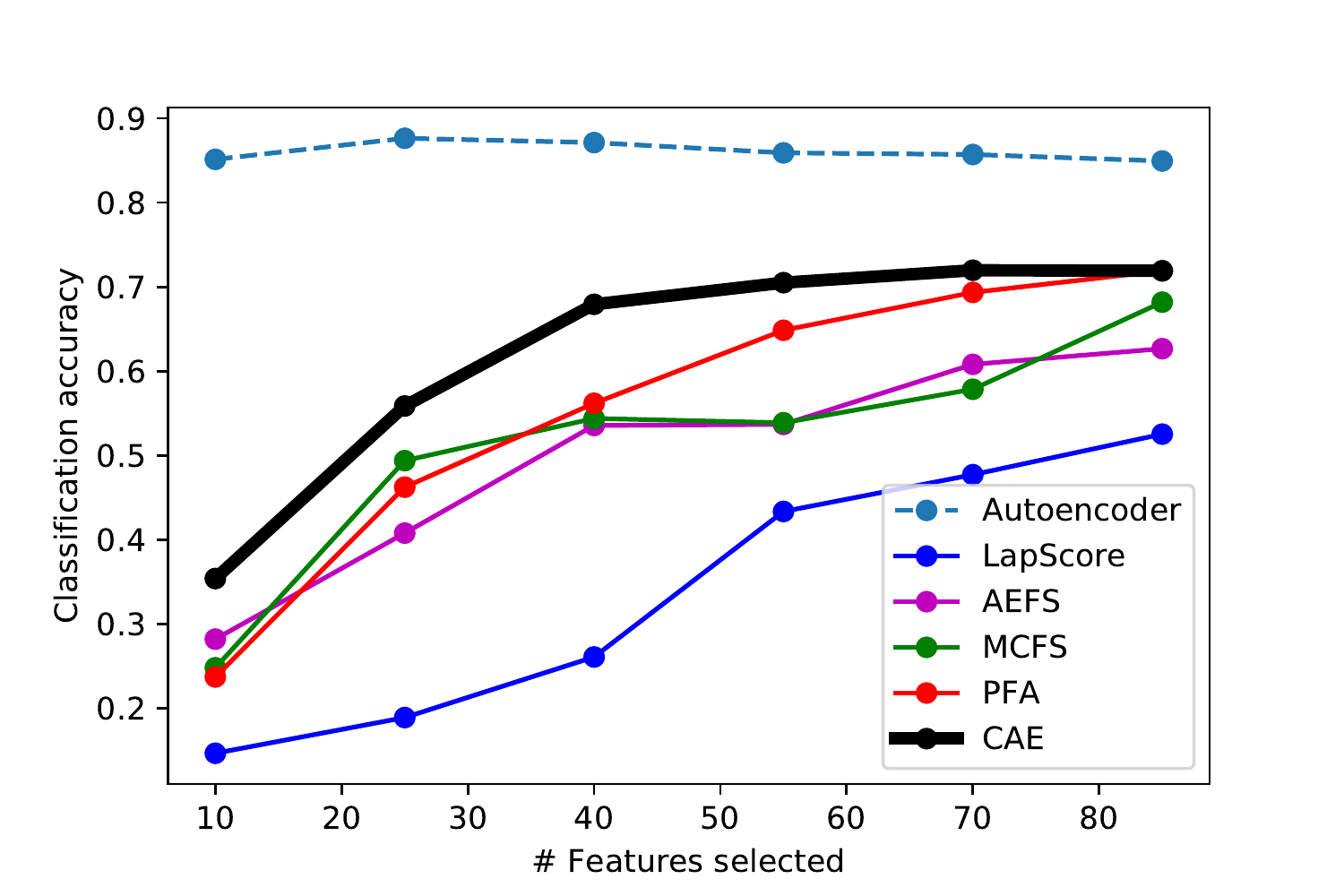}}}
\vspace{-0.3cm}
\end{figure*}

We evaluate these methods on a number of datasets (the sizes of the datasets can be found in Table \ref{tab:1}): 

\textbf{MNIST and MNIST-Fashion} consist of 28-by-28 grayscale images of hand-written digits and clothing items, respectively. We choose these datasets because they are widely known in the machine learning community. Although these are image datasets, the objects in each image are centered, which means we can meaningfully treat each 784 pixels in the image as a separate feature.

\textbf{ISOLET} consists of preprocessed speech data of people speaking the names of the letters in the English alphabet. This dataset is widely used as a benchmark in the feature selection literature. Each feature is one of the 617 quantities produced as a result of preprocessing, including spectral coefficients and sonorant features. 

\textbf{COIL-20} consists of centered grayscale images of 20 objects. Images of the objects were taken at pose intervals of 5 degrees amounting to 72 images for each object. During preprocessing, the images were resized to produce 20-by-20 images, with each feature being one of the 400 pixels.

\textbf{Smartphone Dataset for Human Activity Recognition} consists of sensor data collected from a smartphone mounted on subjects while they performed several activities such as walking upstairs, standing and laying. Each feature represents one of the 561 raw or processed quantities from the sensors on the phone.

\textbf{Mice Protein Dataset} consists of protein expression levels measured in the cortex of normal and trisomic mice who had been exposed to different experimental conditions. Each feature is the expression level of one protein.  

\textbf{GEO Dataset} consists of gene expression profiles measured from a large number of samples through a microarray-based platform. Each of the 10,463 features represents the expression level of one gene.

To evaluate the various feature selection methods, we examine two metrics, both reported on a hold-out test set:

\textbf{Reconstruction error}: We extract the $k$ selected features. We pass the resulting matrix $X_S$ through the reconstruction function $f_\theta$ that we have trained. We measure the Frobenius norm between the original and reconstructed test matrices $\lVert f_\theta(X_S) - X \lVert_F$, normalized by the number of features $d$.
    
 \textbf{Classification accuracy}: We extract the $k$ features selected by the method. We then pass the resulting matrix $X_S$ to an extremely randomized trees classifier \citep{geurts2006extremely}, a variant of random forests that has been used with feature selection methods in prior literature \citep{drotar2015experimental}. We measure the accuracy between the predicted labels and true labels, which are available for each of the datasets. Note that the labels are only used for training the classifier and not for the feature selection.


\subsection{Concrete Autoencoders (Non-Linear Decoder)}

First, we constructed a concrete autoencoder with a non-linear decoder architecture consisting of one hidden layer with $3k/2$ neurons, with $k$ being the number of selected features. We performed a series of experiments with the ISOLET dataset, which is widely used as a benchmark in prior feature selection literature. We benchmarked each feature selection method (besides UDFS whose run-time was prohibitive) with varying numbers of of features (from $k=10$ to $k=85$), measuring the reconstruction error using a 1-hidden-layer neural network as well as classification accuracy. The number of neurons in the hidden layer of the reconstruction network was varied within $[4k/9, 2k/3, k, 3k/2]$, and the network with the highest validation accuracy was selected and measured on the test set.

To control for the performance of the reconstruction network, we trained each reconstruction network for the same number of epochs, 200. For the concrete autoencoder, we did not use the decoder that was learned during training, but re-trained the reconstruction networks from scratch. Our resulting classification accuracies and reconstruction error on each dataset are shown in Fig. \ref{fig:results_nonlinear}. We find that the concrete autoencoder consistently outperformed other feature selection methods on the ISOLET dataset.

\subsection{Concrete Autoencoders (Linear Decoder)}
\label{section:experiments_nonlinear}

Next, we carried out a series of experiments in which we compared concrete autoencoders with linear decoders to the other methods using linear regression as the reconstruction function. Since linear regression can be trained easily to convergence, this allowed us to isolate the effect of using the concrete selector layer for feature selection, and allowed us to train on a wider variety of datasets with less risk of overfitting. We selected a fixed number $k=50$ of features with each method, with the exception of the Mice Protein Dataset, for which we used $k=10$ due to its small size.

After selecting the features using concrete autoencoder and the other feature selection methods, we trained a standard linear regressor with no regularization to impute the original features. The resulting reconstruction errors on a hold-out test set are shown in Table \ref{tab:1}. We also used the selected features to measure classification accuracies, which are shown in Table \ref{tab:2} in Appendix \ref{appendix:classification_accuracies}. On almost all datasets, we found that the concrete autoencoder continued to have the lowest reconstruction error and a high classification accuracy.

\begin{table*}[t]
\begin{center}
 \begin{tabular}{| m{2.3cm} | m{2.2cm} | m{1.2cm} | m{1cm} | m{1cm} | m{1cm} | m{1.3cm} | m{1.6cm} |  m{1.6cm} |}
 \hline
 Dataset & $(n, d)$ & \textit{PCA} & Lap & AEFS & UDFS & MCFS & PFA & CAE \\ 
 \hline
 \hline
 MNIST & (10000, 784) & \textit{0.012} & 0.070 & 0.033 & 0.035 & 0.064 & 0.051 & \textbf{0.026} \\ 
 \hline
 MNIST-Fashion & (10000, 784) & \textit{0.012} & 0.128 & 0.047 & 0.133 & 0.096 & 0.043 & \textbf{0.041} \\ 
 \hline
 COIL-20 & (1440, 400) & \textit{0.008} & 0.126 & 0.061 & 0.116 & 0.085 & \textbf{0.061} & 0.093 \\ 
 \hline
 Mice Protein & (1080, 77) & \textit{0.003} & 0.603 & 0.783 & 0.867 & 0.695 & 0.871 & \textbf{0.372}\\ 
 \hline
 ISOLET & (7797, 617) & \textit{0.009} & 0.344 & 0.301 & 0.375 & 0.471 & 0.316 & \textbf{0.299}\\ 
 \hline
 Activity & (5744, 561) & \textit{0.002} & 0.139 & 0.112 & 0.173 & 0.170 & 0.197 & \textbf{0.108} \\  
 \hline
\end{tabular}
\caption{\textbf{Reconstruction errors of feature selection methods using linear regression reconstruction.} Here, we show the mean-squared errors of the various feature methods on six publicly available datasets. Here Lap refers to the Laplacian score and CAE refers to the concrete autoencoder. For each method, we select $k=50$ features (except for mice protein, where we use $k=10$ because the original data is lower dimensional) and use a linear regressor for reconstruction. All reported values are on a hold-out test set. (Lower is better.) } \label{tab:1}
\end{center}
\end{table*}

\subsection{Interpreting Related Features}
\label{section:experiments_interpretation}

An added benefit of using the concrete selector layer is that it allows the user to not only identify the most informative features for reconstruction, but also identify sets of related features through examination of the learned Concrete parameters $\alphab^{(i)}$. Because the concrete selector layer samples the input features stochastically based on $\alphab^{(i)}$, any of the features with the large values in the vector $\alphab^{(i)}$ may be selected, and are thus likely to be correlated to one another.

\begin{figure}[!htb]
\centering
\subfloat{\includegraphics[width=0.4\linewidth]{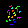}}
\vspace{-0.3cm}
\caption{\textbf{Pixel groups selected by concrete selector nodes on MNIST.} Here, we illustrate the top 3 pixels selected by each of the 20 nodes in the concrete layer when trained on the MNIST dataset. We color each group of 3 pixels with the same color (note that some colors are repeated because of the limited color palette). \textit{cf.} Appendix \ref{appendix:pxel_groups}, which shows pixel group for classes of digits.} \label{fig:selected_pixel_groups}
\end{figure}

In Fig. \ref{fig:selected_pixel_groups}, we show how this can reveal related features by visualizing the top 3 pixels with the highest values in the $\alphab^{(i)}$ vector for each of the 20 concrete selector nodes on the MNIST digits. We notice that the pixels that are selected by each node are spatially close to one another, which agrees with intuitive notions of related features, as neighboring pixel values are likely to be correlated in handwritten digits. These patterns are even more striking when generated for individual classes of digits; in that case, the set of correlated pixels may even suggest the direction of the stroke when the digit was written (see Appendix \ref{appendix:pxel_groups} for more details). Such analysis may be carried out more generally to find sets of related features, such as sets of related genes in a gene expression dataset.

\subsection{Case Study: L1000 Gene Expression}
\label{section:l1000}

We now turn to a large-scale test of the concrete autoencoder: examining whether we can improve \textit{gene expression inference}. Gene expression inference arises from an important problem in molecular biology: characterizing the state of cells in different biological conditions. In particular, the response of cells to diseases, mutations, and drugs is often characterized by the measurement of gene expression patterns \citep{lamb2006connectivity}.

However, measuring all of the genes expressed in a human cell can be expensive, and thus researchers have looked to computational methods to reduce the cost and time required for gene expression profiling. In particular, researchers from the LINCS Project found that, because gene expression is correlated in different conditions, a set of roughly a thousand carefully-chosen genes can capture most of the gene expression information in the entire human transcriptome \citep{peck2006method}. It is thus possible to use a linear regression model trained on genome-wide gene expression to \textit{infer} the gene expression values of the remaining genes. More recently, \citet{chen2016gene} showed that it is possible to leverage the representation power of neural networks to improve the accuracy of gene expression inference in an approach they referred to as D-GEX.

Here, we ask whether it is possible to use concrete autoencoders to determine a good subset of genes, perhaps as an alternative to the landmark genes, without utilizing any prior biological knowledge of gene networks or gene function. We relied only on a large dataset of gene expression data, from which we aim to select the most informative features.

We used the version of the GEO dataset used in the D-GEX paper, and followed the same preprocessing scheme to obtain a dataset of sample size 112,171 and dimensionality 10,463 genes. We then randomly partitioned the dataset in a manner similar to that performed by \citet{chen2016gene}: as a result, the training set had 88,807 samples, the validation set had 11,101, and the test set had 12,263. We then considered 3 kinds of reconstruction functions: in the simplest case, we considered multitarget linear regression, and we also implemented neural networks with 1  and 2 hidden layers. See Appendix \ref{appendix:geo} for the architecture of the networks.

First, we trained a concrete autoencoder to select 943 using only a linear regression decoder. An analysis of the selected genes showed very little overlap with the landmark genes: only 90 of the CAE-selected 943 genes were among the landmark genes. For consistency with the D-GEX results, we used this same set of 943 genes, selected by a concrete autoencoder with a linear decoder, with all of our reconstruction networks. 

\begin{figure}[!htb]
\centering
\subfloat[]{\includegraphics[width=0.49\linewidth]{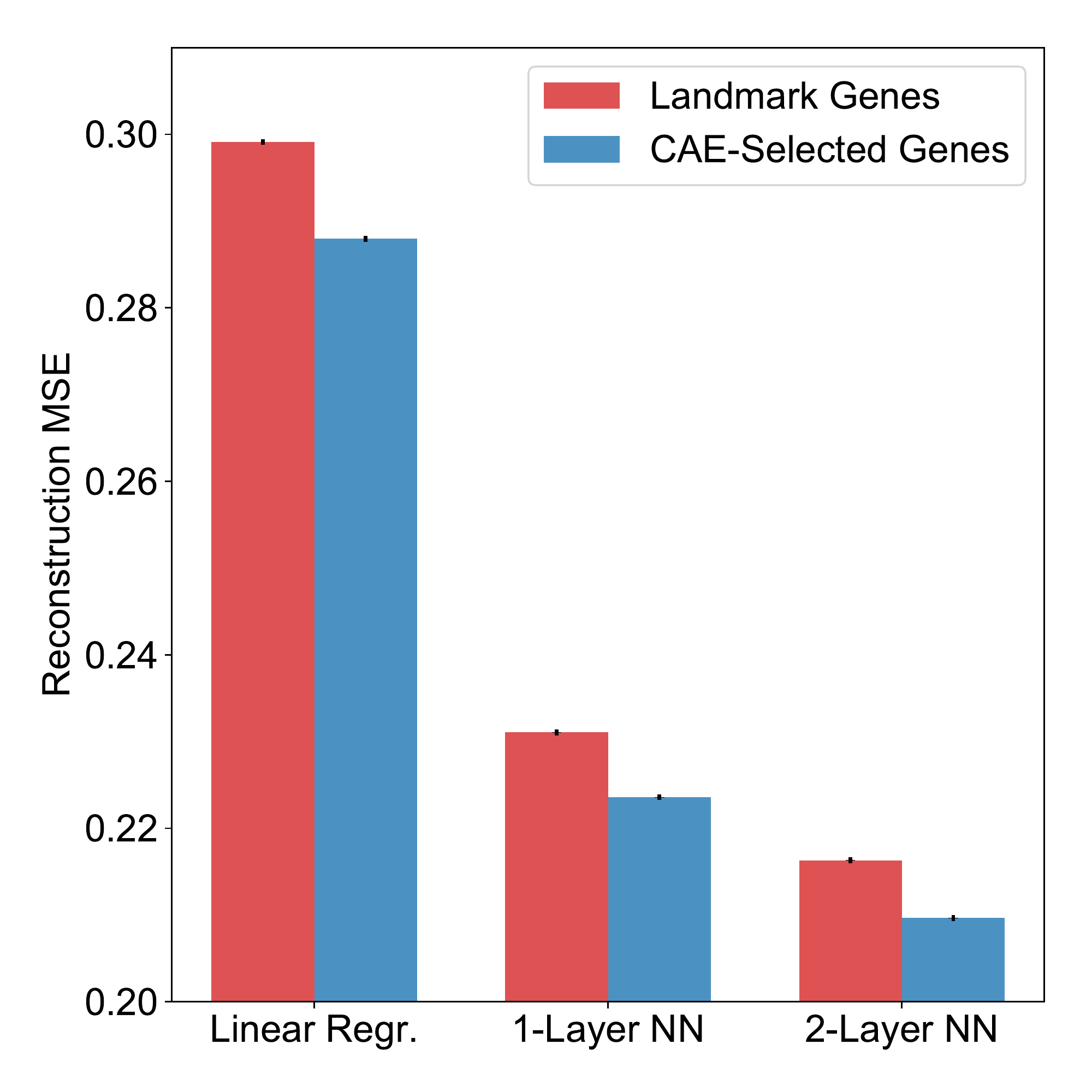}} 
\subfloat[]{\includegraphics[width=0.49\linewidth]{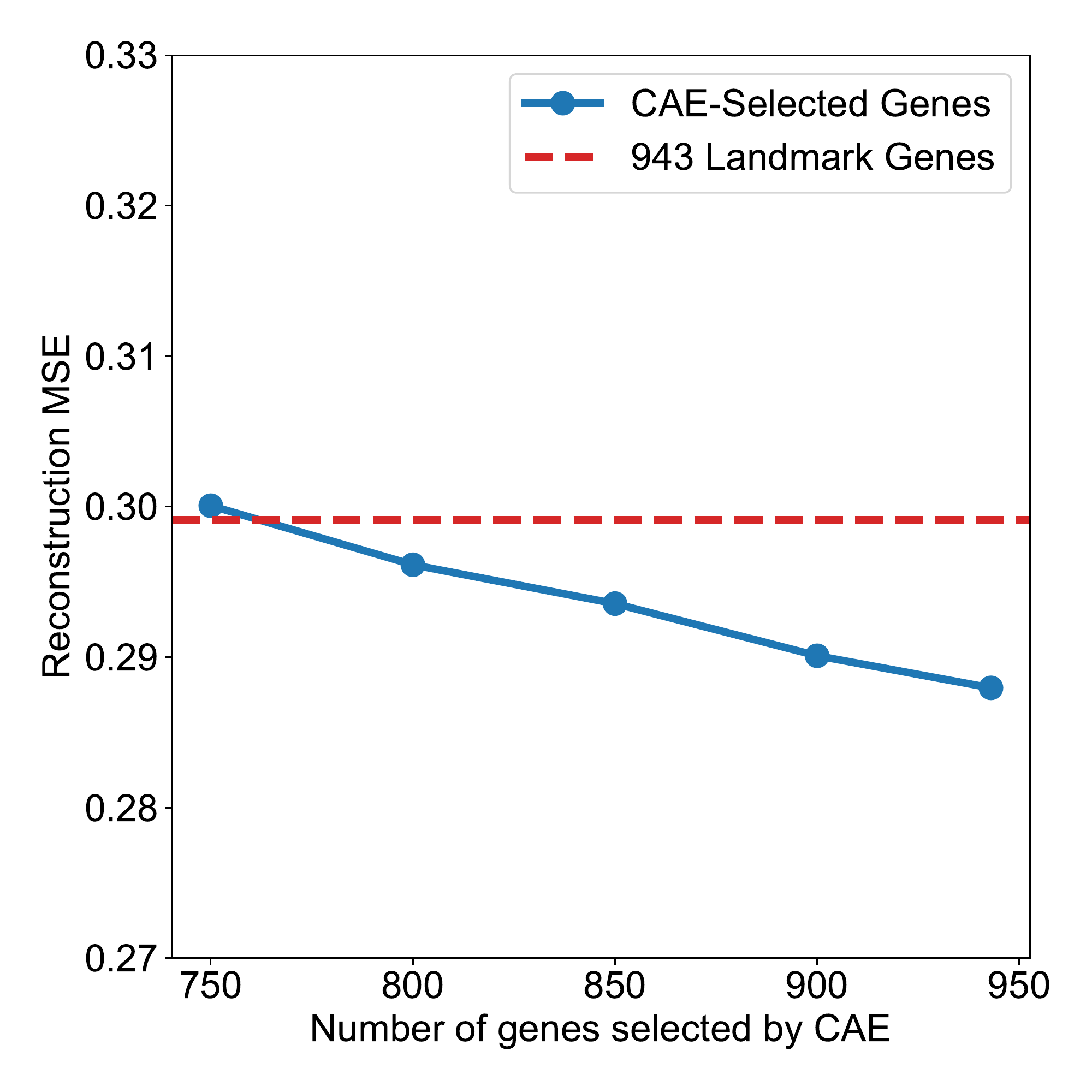}} 
\vspace{-0.3cm}
\caption{\textbf{Imputation errors of concrete autoencoders and landmark genes.} Here, we show the mean-squared error of the imputation task using both the 943 landmark genes (\textcolor{red}{red}) and the 943 genes selected by the concrete autoencoder (\textcolor{blue}{blue}) on the test set. The task is to impute the expression of all 10,463 genes.  We observe about a 3\% reduction (note that y-axis begins at 0.20) of the reconstruction error when using the  genes selected by the concrete autoencoders (CAE) across all architectures. These are results averaged over three trials. Standard deviation bars are shown but were very low, as the final imputations were very similar across all trials. (b) We train the CAE with different numbers of selected features, and calculate the MSE using linear regression on the test set. We find that we can achieve a similar MSE to the landmark genes using only around 750 genes, a 20\% reduction in the number of genes measured.} \label{fig:l1000}
\end{figure}

We trained each reconstruction networks to impute all of the original 10,463 genes. We measured the reconstruction error on a hold-out test set that was used neither to train the concrete autoencoder nor the reconstruction functions. Our results are summarized in Fig. \ref{fig:l1000}(a), where we plot the mean-squared error of imputation. We show that not only is it possible to use concrete autoencoders to perform gene selection gene expression inference in a differentiable, end-to-end manner on large-scale datasets, doing so improves the performance of the gene expression imputation on a holdout test set of gene expression by around 3\% for each architecture, which is significant as the L1000 landmark genes were expert curated using a combination of computational prediction with domain knowledge, and is a very strong benchmark and is widely used in genomics. 

Next, we investigated whether it would be possible to obtain similar accuracies as to the landmark genes while using a smaller set of CAE-selected genes. We trained concrete autoencoders from scratch using $k=750, 800, 850, 900, 943$, using the same architecture described in Appendix \ref{appendix:geo} and using a linear regression decoder. We found that using linear regression as the reconstruction function, we could obtain reconstruction MSEs about as low as the landmark genes, using only 750 genes, which represents roughly a $20 \%$ reduction in the number of genes measured, potentially saving substantial experimental costs. These results are illustrated in Fig. \ref{fig:l1000}(b).

\section{Discussion}

In this paper, we have proposed a new method for differentiable, end-to-end feature selection via backpropagation. At its core, the concrete autoencoder uses Concrete random variables and the reparametrization trick to allow gradients to flow through a layer that stochastically selects discrete input features. The stochasticity of the concrete autoencoder allows it to efficiently explore and converge to a subset of input features of specified size that minimizes a particular loss, as described in Section \ref{section:method}. The learned parameters can be further probed to allow the anlayst to interpret related features, as demonstrated in Section \ref{section:experiments_interpretation}. This makes concrete autoencoders different from many competing methods, which rely on regularization to encourage sparse feature selection.

We show via experiments on a variety of public datasets that concrete autoencoders effectively minimize the reconstruction error and maximize classification accuracy using selected features. In the six public datasets that we tested concrete autoencoders, we found that concrete autoencoders outperformed many different complex feature selection methods. This remains the case even when we reduced the decoder layer to be a single linear layer, showing that the concrete selector node is useful even when selecting input features that minimize the loss when using a linear regression as the reconstruction function. 

Because the concrete autoencoder is an adaptation of the standard autoencoder, it  scales easily to datasets with many samples or high dimensionality. We demonstrated this in section \ref{section:l1000} using a gene expression dataset with more than 100,000 samples and 10,000 features, where the  features selected by the concrete autoencoder outperformed the state-of-the-art gene subset. Furthermore, because of its general formulation, the concrete autoencoder can be  easily extended in many ways.  For example, it possible to use concrete autoencoders in a supervised manner -- to select a set of features that minimize a cross-entropy loss, for example, rather than a reconstruction loss. More details and examples of this approach are provided in Appendix \ref{appendix:supervised}. Another possible extension is to attach different costs to selecting different features, for example if certain features represent tests or assays that are much more expensive than others. Such as cost may be incorporated into the loss function and allow the analyst to trade off cost for accuracy.

Advantages of the concrete autoencoder include its generality and ease of use. Implementing the architecture in popular machine learning frameworks requires only modifying a few lines of code from a standard autoencoder. Furthermore, the runtime of the concrete autoencoder is similar to that of the standard autoencoder and improves with hardware acceleration and parallelization techniques commonplace in deep learning. The only additional hyperparameters of the concrete autoencoder are the initial and final temperatures used in the annealing schedule. We find that the default values used in this paper work well for a variety of datasets. 

Concrete autoencoders, like the other feature selection methods we compared with in this paper, do not provide \textit{p}-values or statistical significance quantification. Features discovered through concrete autoencoders should be validated through hypothesis testing or
additional analysis using relevant domain knowledge. We believe that the concrete autoencoder can be of particular use in simplifying assays and experiments that measure a large number of related quantities, such as medical lab tests and genotype sequencing. 

\label{section:discussion}

\clearpage

\section*{Acknowledgments}

We are grateful for helpful comments by Amirata Ghorbani in the development of this technique, as well to Ali Abid and Aneeqa Abid for feedback regarding figures. The authors also thank Manway Liu, Brielin Brown, Matt Edwards, and Thomas Snyder for discussions that inspired this project.

\bibliography{main}
\bibliographystyle{icml2019}

\appendix
\newpage
\begin{onecolumn}

\section*{Appendices}
\section{Selected Features for Single Classes in MNIST}
\label{appendix:single-classes}

Here, we show additional examples of using the concrete autoencoder on subsets of the MNIST data that consist of a single digit. Here, we select $k=10$ features for each subset of data.

\begin{figure*}[!htb]
\centering
\subfloat[]{\includegraphics[width=0.24\linewidth]{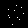}} \;
\subfloat[]{\includegraphics[width=0.24\linewidth]{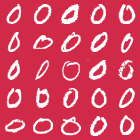}} \;
\subfloat[]{\includegraphics[width=0.24\linewidth]{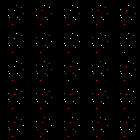}} \;
\subfloat[]{\includegraphics[width=0.24\linewidth]{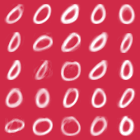}} 
\vspace{-0.3cm}
\caption{Here, we show the results of using concrete autoencoders to select the $k=10$ most informative pixels of images of the digit 0 in the MNIST dataset. Compare with Fig. \ref{fig:mnist-full} in the main paper for more information.} \label{fig:mnist-0}
\end{figure*}

\begin{figure*}[!htb]
\centering
\subfloat[]{\includegraphics[width=0.24\linewidth]{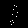}} \;
\subfloat[]{\includegraphics[width=0.24\linewidth]{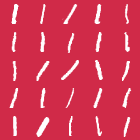}} \;
\subfloat[]{\includegraphics[width=0.24\linewidth]{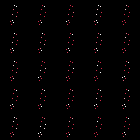}} \;
\subfloat[]{\includegraphics[width=0.24\linewidth]{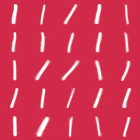}} 
\vspace{-0.3cm}
\caption{Here, we show the results of using concrete autoencoders to select the $k=10$ most informative pixels of images of the digit 0 in the MNIST dataset. Compare with Fig. \ref{fig:mnist-full} in the main paper for more information.} \label{fig:mnist-0}
\end{figure*}

\begin{figure*}[!htb]
\centering
\subfloat[]{\includegraphics[width=0.24\linewidth]{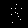}} \;
\subfloat[]{\includegraphics[width=0.24\linewidth]{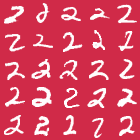}} \;
\subfloat[]{\includegraphics[width=0.24\linewidth]{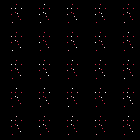}} \;
\subfloat[]{\includegraphics[width=0.24\linewidth]{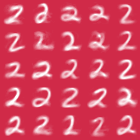}} 
\vspace{-0.3cm}
\caption{Here, we show the results of using concrete autoencoders to select the $k=10$ most informative pixels of images of the digit 0 in the MNIST dataset. Compare with Fig. \ref{fig:mnist-full} in the main paper for more information.} \label{fig:mnist-0}
\end{figure*}

\section{Pseudocode for a Concrete Autoencoder}
Here, we have the pseudocode for training the concrete autoencoder in more detail. We also describe how to use a trained concrete autoencoder for feature selection on new data, as well as how to use the concrete autoencoder for imputation.

\label{appendix:pseudocode}



  

\begin{algorithm}[H]
  \caption{Training a Concrete Autoencoder}
  \label{alg:iterative}
  \begin{algorithmic}
  \STATE {\bfseries Input:} training dataset $X \in \mathbb{R}^{n \times d}$, number of features to select $k$, decoder network $f_\theta(\cdot)$, number of epochs $B$, learning rate $\lambda$, initial temp $T_0$, final temp $T_B$.
  \STATE 
  \FOR {$i \in \{1 \ldots k\}$} 
    \STATE Initialize a $d$-dimensional vector of parameters $\alphab^{(i)}$ with small positive values.
\ENDFOR
    \STATE Initialize the parameters $\theta$ of the reconstruction function in a standard way for neural networks.  
  \FOR {$b \in \{1 \ldots B\}$} 
  \STATE Let $T = T_0 (T_B / T_0) ^{b/B}$
  \FOR {$i \in \{1 \ldots k\}$} 
    \STATE Sample $\mb^{(i)} \sim  $ Concrete$(\alphab^{(i)}, T)$
    \STATE {Let $X^{(i)}_S = X \cdot \mb^{(i)}$}

  \ENDFOR
    \STATE Define $X_S$ to be the matrix $\in \mathbb{R}^{n \times k}$ that results from horizontally concatenating the $X^{(1)}_S \ldots X^{(k)}_S$.
    \STATE {Let the loss $L$ be defined as $ \lVert f_\theta(X_S) - X \lVert_F$}
    \STATE Compute the gradient of the loss w.r.t. $\theta$ using backpropagation and w.r.t each $\alphab^{(i)}$ using the reparametrization trick.
    \STATE Update the parameters $\theta \leftarrow \theta - \lambda \nabla_\theta L $, and $\alphab^{(i)} \leftarrow \alphab^{(i)} - \lambda \nabla_{\alphab^{(i)}} L $ 
  \ENDFOR
  \STATE
  \STATE {\bfseries Return:} trained reconstruction function $f_\theta(\cdot)$ and trained Concrete parameters $\alphab^{(i)}$
  
  \end{algorithmic}
\end{algorithm}

\begin{algorithm}[H]
  \caption{Using a Trained Concrete Autoencoder for Feature Selection}
  \label{alg:iterative}
  \begin{algorithmic}
  \STATE {\bfseries Input:} test sample $\xb \in \mathbb{R}^{d}$, trained Concrete parameters $\alphab^{(i)}$
  \STATE
 \FOR {$i \in \{1 \ldots k\}$} 
    \STATE {Let $m^{(i)} = \argmax_j(\alphab^{(i)}_j)$, where $j$ indexes the elements of the sample vector}
    \STATE {Let $\xb_S^{(i)} = \xb_{m^{(i)}}$}    
 \ENDFOR
 \STATE 
 \STATE {\bfseries Return: $\xb_S$}
  \end{algorithmic}
\end{algorithm}

\begin{algorithm}[H]
  \caption{Using a Trained Concrete Autoencoder for Imputation}
  \label{alg:iterative}
  \begin{algorithmic}
  \STATE {\bfseries Input:} test sample with subset of features $\hat{\xb} \in \mathbb{R}^{k}$, trained reconstruction function $f_\theta(\cdot)$
  \STATE 
 \STATE {\bfseries Return: $f_\theta(\hat{\xb})$}
  \end{algorithmic}
\end{algorithm}


\clearpage

\section{Classification Accuracies for Feature Selection Methods with Linear Reconstruction}
\label{appendix:classification_accuracies}

We carried out a series of experiments in which we compared concrete autoencoders with linear decoders to the other feature selection methods using linear regression as the reconstruction function. We selected $k=50$ of features with each method.

After selecting the features using concrete autoencoder and the other feature selection methods, we trained a standard linear regressor with no regularization to impute the original features. The resulting reconstruction errors on a hold-out test set are shown in Table \ref{tab:1} in the main text. We also used the selected features to measure classification accuracies, which are shown in Table \ref{tab:2} here. Generally, we find that the concrete autoencoder continues to have the lowest reconstruction error and a high (but not always the highest) classification accuracy.

\begin{table}[H]
\begin{center}
 \begin{tabular}{| m{2.3cm} | m{2.2cm} | m{1.2cm} | m{1cm} | m{1cm} | m{1cm} | m{1.3 cm} | m{1.6 cm} | m{1.6 cm} |}
 \hline
 Dataset & $(n, d)$ & \textit{PCA} & Lap & AEFS & UDFS & MCFS & PFA & CAE \\ 
 \hline
 \hline
 MNIST & (10000, 784) & \textit{0.925} & 0.646 & 0.690 & 0.892 & 0.807 & 0.852 & \textbf{0.906} \\ 
 \hline
 MNIST-Fashion & (10000, 784) & \textit{0.825} & 0.517 & 0.580 & 0.547 & 0.513 & \textbf{0.683} & 0.677 \\ 
 \hline
 COIL-20 & (1440, 400) & \textit{0.996} & 0.389 & 0.580 & 0.556 & 0.635 & \textbf{0.642} & 0.586 \\ 
 \hline
 Mice Protein & (1080, 77) & \textit{0.721} & 0.134 & 0.125 & \textbf{0.139} & \textbf{0.139} & 0.130 & 0.134 \\
 \hline
 ISOLET & (7797, 617) & \textit{0.895} & 0.407 & 0.576 & 0.455 & 0.522 & 0.622 & \textbf{0.685} \\ 
 \hline
 Activity & (5744, 561) & \textit{0.796} & 0.280 & 0.240 & 0.287 & 0.295 & 0.364 & \textbf{0.420} \\  
 \hline
\end{tabular}
\caption{\textbf{Classification accuracies of feature selection methods.} Here, we show the classification accuracies of the various feature methods on six publicly available datasets. Here CAE refers to the concrete autoencoder. For each method, we select $k=50$ features (except for mice protein dataset, for which we use $k=10$) and use a neural network with 1 hidden layer for reconstruction. All reported values are on the test set. The classifier used here was a Extremely Randomized Trees classifier (a variant of Random Forests) with the number of trees being 50. (Higher is better.)} \label{tab:2}
\end{center}
\end{table}

\clearpage

\section{Examples of Feature Groups in MNIST Digits}
\label{appendix:pxel_groups}

Here, we show examples of feature groups that were selected by the concrete autoencoder on single classes of digits in the MNIST dataset (see Section \ref{section:experiments_interpretation} in the main text for more details). The patterns in the case of single classes of digits are even more striking; here, the set of correlated pixels can be used to infer the direction of the stroke when the digit was written, as correlated pixels are more likely to be part of the same stroke. For example, consider Fig. \ref{fig:pixel-groups-digits}(b), in which the pixel groups shown for the digit `1'. We note that the pixel groups tend to form vertical subsets, as the writing stroke connected those sets of pixels together.

\begin{figure*}[!bth]
\centering
\subfloat[]{\includegraphics[width=0.21\linewidth,]{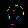}} \;
\subfloat[]{\includegraphics[width=0.21\linewidth,]{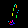}} \;
\subfloat[]{\includegraphics[width=0.21\linewidth,]{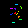}} \; \;
\subfloat[]{\includegraphics[width=0.21\linewidth,]{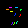}}
\vspace{-0.3cm}
\caption{Here, we show the results of using concrete autoencoders to select the $k=10$ most informative pixels groups of images of the digit 0 in the MNIST dataset. Compare with Fig. \ref{fig:selected_pixel_groups} in the main paper. Each panel is a different digit: (a) the digit 0, (a) the digit 1, (a) the digit 2, (a) the digit 7} \label{fig:pixel-groups-digits}
\end{figure*}

\clearpage

\section{Architecture for GEO Dataset Experiments}
\label{appendix:geo}

For the GEO dataset, we trained three reconstruction networks to perform the imputation from both the landmark and CAE-selected genes. In each case, the hidden layers consisted of 9000 neurons and dropout rate was set to $0.1$. The initial learning rate was $0.001$ and decayed after each epoch by multiplication with $0.95$. The number of epochs was set to 100 for training the reconstruction networks. We used batch sizes of 256.

The concrete autoencoder was trained with a linear decoder network for 5000 epochs. For this dataset, we set the initial temperature to be 10, and the final temperature to be 0.01.  

\clearpage

\section{Supervised Concrete Autoencoders}
Concrete autoencoders can be easily adapted to the supervised setting by replacing the reconstruction neural network in the decoder with a neural network classifier. The pseudocode, shown below, is quite similar to training the standard concrete autoencoder.

\begin{algorithm}[H]
  \caption{Training a Concrete Autoencoder}
  \label{alg:iterative}
  \begin{algorithmic}
  \STATE {\bfseries Input:} training features $X \in \mathbb{R}^{n \times d}$, training labels $\yb$, number of features to select $k$, classifier function $f_\theta(\cdot)$, number of epochs $B$, learning rate $\lambda$, initial temperature $T_0$, final temperature $T_B$.
  \STATE 
  \FOR {$i \in \{1 \ldots k\}$} 
    \STATE Initialize a $d$-dimensional vector of parameters $\alphab^{(i)}$ with small positive values.
\ENDFOR
    \STATE Initialize the parameters $\theta$ of the reconstruction function in a standard way for neural networks.  
  \FOR {$b \in \{1 \ldots B\}$} 
  \STATE Let $T = T_0 (T_B / T_0) ^{b/B}$
  \FOR {$i \in \{1 \ldots k\}$} 
    \STATE Sample $\mb^{(i)} \sim  $ Concrete$(\alphab^{(i)}, T)$
    \STATE {Let $X^{(i)}_S = X \cdot \mb^{(i)}$}

  \ENDFOR
    \STATE Define $X_S$ to be the matrix $\in \mathbb{R}^{n \times k}$ that results from horizontally concatenating the $X^{(1)}_S \ldots X^{(k)}_S$.
    \STATE {Let $L$ be the cross entropy loss between the true labels $\yb$ and the logits $f_\theta(X_S)$}
    \STATE Compute the gradient of the loss w.r.t. $\theta$ using backpropagation and w.r.t each $\alphab^{(i)}$ using the reparametrization trick.
    \STATE Update the parameters $\theta \leftarrow \theta - \lambda \nabla_\theta L $, and $\alphab^{(i)} \leftarrow \alphab^{(i)} - \lambda \nabla_{\alphab^{(i)}} L $ 
  \ENDFOR
  \STATE
  \STATE {\bfseries Return:} trained reconstruction function $f_\theta(\cdot)$ and trained Concrete parameters $\alphab^{(i)}$
  
  \end{algorithmic}
\end{algorithm}

We trained a concrete autoencoder in this supervised manner on the MNIST digits, some representative images are shown in Fig. \ref{fig:mnist-full-supervised}. Generally, we found the imputation quality to be not as good as when the objective function is directly reconstruction error.

\label{appendix:supervised}

\begin{figure*}[!bth]
\centering
\subfloat[]{\includegraphics[width=0.24\linewidth]{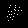}} \;
\subfloat[]{\includegraphics[width=0.24\linewidth]{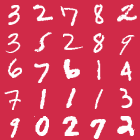}} \;
\subfloat[]{\includegraphics[width=0.24\linewidth]{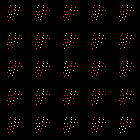}} \;
\subfloat[]{\includegraphics[width=0.24\linewidth]{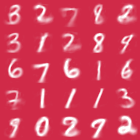}} 
\vspace{-0.3cm}
\caption{\textbf{Demonstrating concrete autoencoders on the MNIST dataset.} Here, we show the results of using concrete autoencoders to select in an supervised manner the $k=20$ most informative pixels of images in the MNIST dataset. (a) The 20 selected features (out of the 784 pixels) on the MNIST dataset are shown in white. (b) A sample of input images in MNIST dataset with the top 2 rows being training images and the bottom 3 rows being test images. (c) The same input images with only the selected features shown as white dots. (d). The reconstructed versions of the images, using only the 20 selected pixels, shows that generally the digit is identified correctly.} \label{fig:mnist-full-supervised}
\end{figure*}

\end{onecolumn}




\end{document}